\RequirePackage[T1]{fontenc}
\documentclass[letterpaper,10pt,conference]{ieeeconf}
\IEEEoverridecommandlockouts
\usepackage{graphicx}
\usepackage{amsmath,amssymb}
\usepackage{booktabs,array}
\usepackage{cite}
\usepackage{url}
\usepackage[hidelinks]{hyperref}
\let\AGthebibliography\thebibliography
\renewcommand{\thebibliography}[1]{\AGthebibliography{#1}\interlinepenalty=10000}
\newcommand{\PiTape}{5}
\newcommand{\PiSponge}{5}
\newcommand{\PiMaze}{4}
\newcommand{\PiDrawer}{6}
\newcommand{\PiWater}{4}
\newcommand{\PiSlipper}{5}
\newcommand{\PiWaste}{6}
\newcommand{\PiShape}{5}
\newcommand{\PiTotal}{40}
\newcommand{\PiOverall}{50.00}
\newcommand{\PivotTape}{5}
\newcommand{\PivotSponge}{5}
\newcommand{\PivotMaze}{4}
\newcommand{\PivotDrawer}{5}
\newcommand{\PivotWater}{4}
\newcommand{\PivotSlipper}{6}
\newcommand{\PivotWaste}{5}
\newcommand{\PivotShape}{4}
\newcommand{\PivotTotal}{38}
\newcommand{\PivotOverall}{47.50}
\newcommand{\AstraTape}{7}
\newcommand{\AstraSponge}{8}
\newcommand{\AstraMaze}{6}
\newcommand{\AstraDrawer}{7}
\newcommand{\AstraWater}{7}
\newcommand{\AstraSlipper}{8}
\newcommand{\AstraWaste}{8}
\newcommand{\AstraShape}{6}
\newcommand{\AstraTotal}{57}
\newcommand{\AstraOverall}{71.25}
\newcommand{\AblFullTape}{20/30 (66.67\%)}
\newcommand{\AblFullMaze}{19/30 (63.33\%)}
\newcommand{\AblFullWater}{20/30 (66.67\%)}
\newcommand{\AblFullWaste}{25/30 (83.33\%)}
\newcommand{\AblSvtTape}{15/30 (50.00\%)}
\newcommand{\AblSvtMaze}{13/30 (43.33\%)}
\newcommand{\AblSvtWater}{9/30 (30.00\%)}
\newcommand{\AblSvtWaste}{16/30 (53.33\%)}
\newcommand{\AblMvecwTape}{10/30 (33.33\%)}
\newcommand{\AblMvecwMaze}{9/30 (30.00\%)}
\newcommand{\AblMvecwWater}{9/30 (30.00\%)}
\newcommand{\AblMvecwWaste}{13/30 (43.33\%)}
\newcommand{\AblVgtgTape}{2/30 (6.67\%)}
\newcommand{\AblVgtgMaze}{3/30 (10.00\%)}
\newcommand{\AblVgtgWater}{1/30 (3.33\%)}
\newcommand{\AblVgtgWaste}{1/30 (3.33\%)}
\newcommand{\AblRcrmTape}{5/30 (16.67\%)}
\newcommand{\AblRcrmMaze}{6/30 (20.00\%)}
\newcommand{\AblRcrmWater}{1/30 (3.33\%)}
\newcommand{\AblRcrmWaste}{5/30 (16.67\%)}
\newcommand{\AstraEvalCount}{\AstraTotal}
\newcommand{\AstraEvalOverall}{\AstraOverall}

\title{\LARGE\bf AntiGrounding: Executable Robot Trajectories as Visual Prompts\protect\\
for VLM-Guided Manipulation}
\author{Wenbo Li$^{1}$, Yiteng Chen$^{1}$, Wenhao Li$^{1}$, and Qingyao Wu$^{1,\dagger}$\\
{\normalsize $^{1}$School of Software Engineering, South China University of Technology}\\
{\normalsize $^{\dagger}$Corresponding author}}

\begin{document}

\maketitle
\thispagestyle{empty}
\pagestyle{empty}
\raggedbottom
\begin{abstract}
Natural-language instructions specify manipulation goals but leave the robot's motion underdetermined. We present AntiGrounding, a visual action-selection framework built around a dual geometric--visual trajectory interface. Each short trajectory retained after feasibility filtering remains an explicit motion plan and serves as a visual prompt for instruction-conditioned vision--language model (VLM) assessment. Structured multi-view visual question answering (VQA) scores safety, task alignment, efficiency, and physical plausibility. Weighted view fusion aggregates these scores for trajectory selection, while the scores also guide subsequent translational proposals. Separate orientation and gripper controls coordinate physical interaction. Planning proceeds in an initialized digital twin, which validates selected segments before the real robot executes the same waypoint sequences. Across eight real-world manipulation tasks, AntiGrounding with a single GPT-6 Astra evaluator achieves \AstraOverall\% overall success. Under the reported deployment protocol, $\pi_{0.5}$ achieves \PiOverall\%, and a PIVOT-style visual proposal-selection baseline with the same evaluator achieves \PivotOverall\%. Component ablations and evaluator sensitivity characterize trajectory assessment, proposal search, orientation control, and evaluator choice. Performance depends on digital-twin fidelity and physical interaction.
\end{abstract}

\begin{figure*}[t]
\centering
\includegraphics[width=0.95\textwidth]{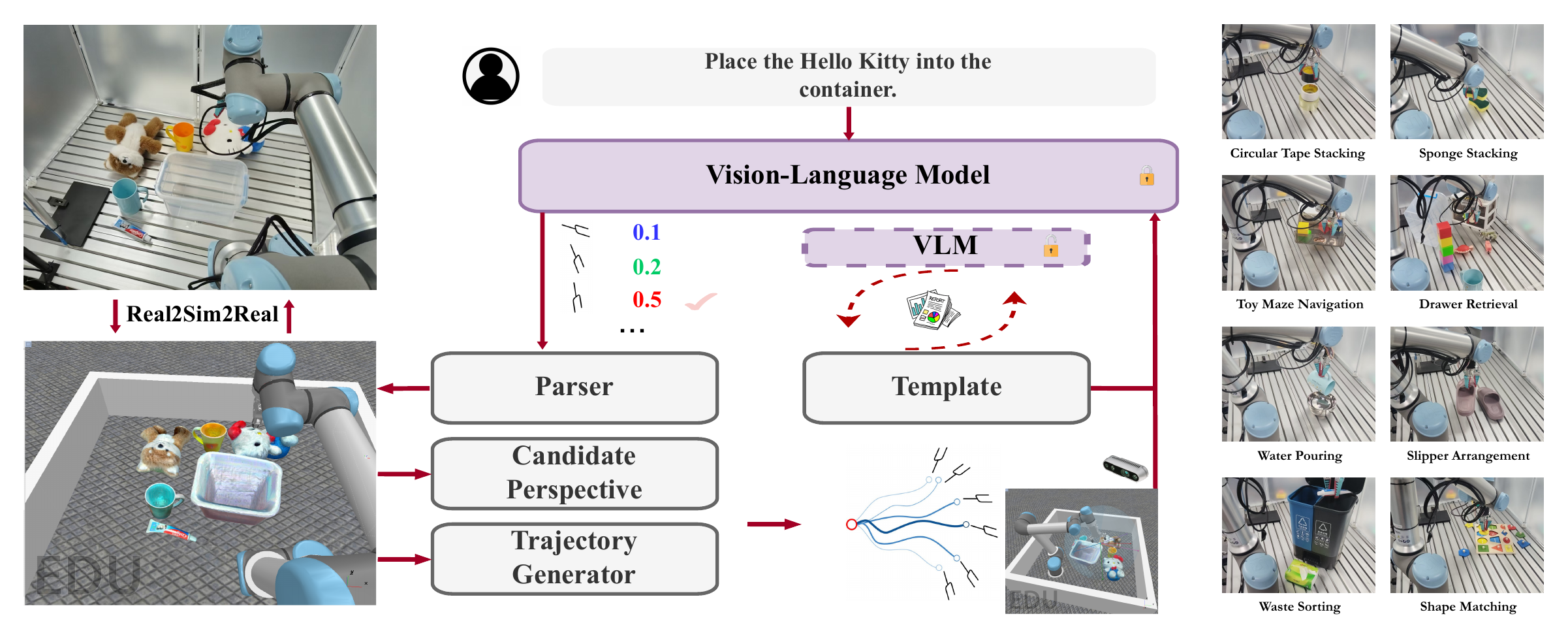}
\caption{\textbf{AntiGrounding overview.}
Trajectories retained after geometric filtering serve as visual prompts for
instruction-conditioned assessment and explicit geometry for validation and
execution. Structured VQA scores the candidates across rendered scene views.
The surrounding panels illustrate the eight manipulation tasks.
Real2Sim2Real comprises initial real-to-sim scene construction followed by
simulator-mediated planning and real execution.}
\label{fig:head}
\end{figure*}

\section{Introduction}
\label{sec:Introduction}
Natural-language instructions specify a manipulation goal rather than a unique robot trajectory. Pouring water into a bowl requires the cup to approach with sufficient clearance, align with the bowl, and rotate at the appropriate task stage. A collision-free path can still miss the bowl, while rotation alone leaves the cup's position unspecified. Vision--language models (VLMs) interpret task semantics and scene context~\cite{saycan,palm-e}, whereas geometric planners represent motion constraints. AntiGrounding connects these task-level judgments to executable motion through trajectory candidates (Fig.~\ref{fig:head}).

The reasoning-to-action interface determines how semantic judgments influence robot control. Code as Policies expresses control through programs and primitives, VoxPoser through spatial value maps, and ReKep through relational keypoint constraints~\cite{codeaspolicy,voxposer,huang2024rekep}. General multimodal models are also beginning to control robots directly or monitor and correct robot-policy execution~\cite{via2026,flare2026}.

PIVOT selects and iteratively refines visually annotated action proposals with VLMs~\cite{pivot}. HAMSTER and MolmoAct likewise demonstrate visual action representations~\cite{hamster,molmoact}. Our focus is the short executable trajectory itself: how to retain it as an explicit geometric control object while exposing it to task-conditioned multimodal assessment. A single projection can obscure alignment relationships, and a holistic judgment can conflate competing assessment criteria.

AntiGrounding samples short translational end-effector trajectories, filters them in simulation, and renders the retained candidates for structured multi-view VQA. View selection identifies useful projections. Criterion weights combine the four assessment dimensions, and retained-view weights fuse their scores. The selected segment undergoes simulator execution validation before the real robot follows the same waypoints. Planning continues from the simulator state, with all candidate scores guiding subsequent proposals. A single GPT-6 Astra model evaluates the rendered candidates.

Our contributions are: (1) a dual geometric--visual interface that maintains each retained trajectory as both an explicit geometric control object and a visual action representation through assessment, simulator validation, and physical execution; (2) structured multi-view trajectory assessment coupled with score-guided proposal refinement for executable action selection; and (3) an eight-task real-world manipulation study, with component ablations and evaluator sensitivity, that characterizes the interface's capabilities and limitations.

\section{Related Work}
\label{sec:RelatedWorks}

\subsection{Action Representations for Manipulation}
Vision--language--action (VLA) models, including OpenVLA, $\pi_0$, and $\pi_{0.5}$, learn instruction-conditioned visuomotor policies~\cite{kim2024openvla,black2024pi_0,intelligence2025pi_0.5}. HAMSTER represents motion through coarse visual paths, while MolmoAct and MolmoAct2 connect embodied reasoning to trajectory traces or continuous actions~\cite{hamster,molmoact,molmoact2}. Programs, spatial value maps, and keypoint constraints provide explicit control structure~\cite{codeaspolicy,voxposer,huang2024rekep}. VIA gives a general-purpose multimodal agent a visual interface for robot control~\cite{via2026}; FLARE combines a $\pi_{0.5}$-based policy with multimodal failure analysis and execution monitoring~\cite{flare2026}. AntiGrounding generates executable candidates geometrically and delegates their task-conditioned assessment to a multimodal evaluator.

\subsection{Visual Action Evaluation}
PIVOT is the closest visual-prompting prior, using iterative VQA to select and refine annotated action proposals~\cite{pivot}. AntiGrounding evaluates short translational trajectories with explicit geometric feasibility, structured multi-view assessment, separate interaction controls, and simulator-validated real execution. VeriSpace learns a spatial action verifier, whereas Uncertainty-aware Policy Steering calibrates VLM verification through conformal prediction~\cite{verispace,ups}. AntiGrounding weights its selected rendered views according to their clarity and usefulness.

\begin{figure*}[t]
\centering
\includegraphics[width=\textwidth]{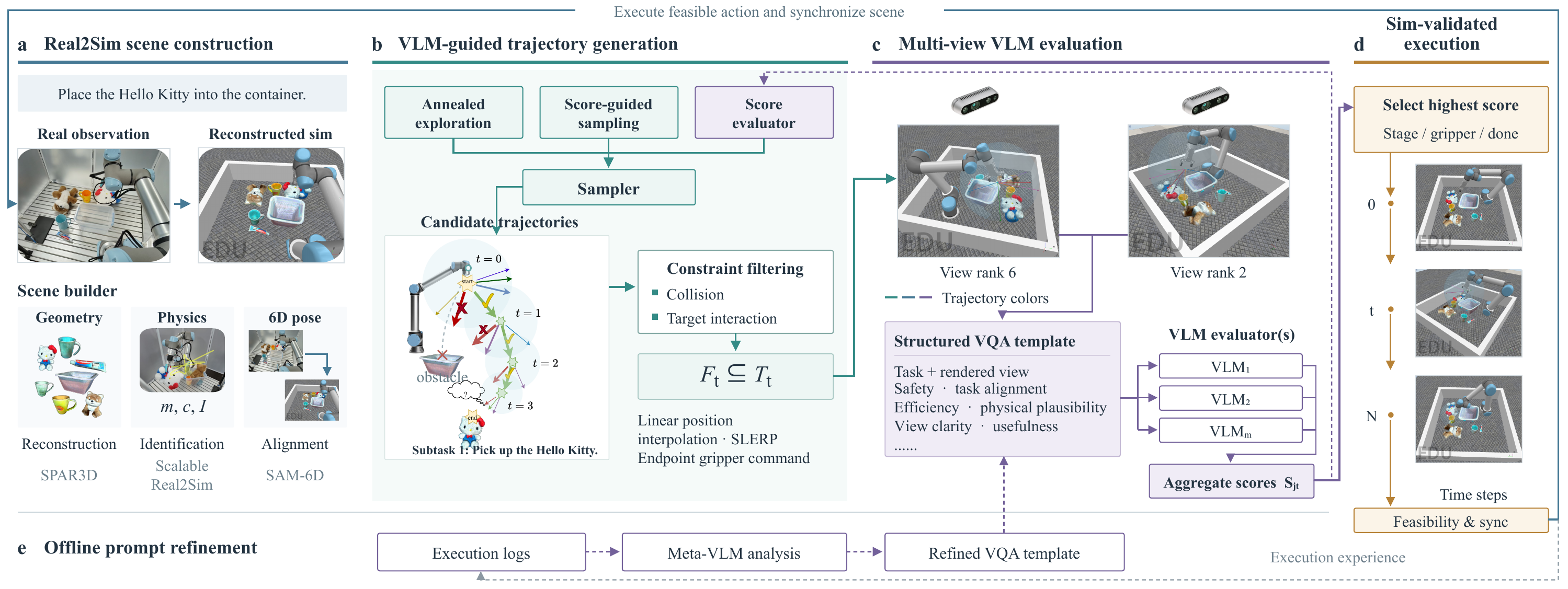}
\caption{\textbf{AntiGrounding framework.} Real2Sim initializes the simulator, which supplies the planning state and rendered VLM observations. \emph{Feasibility \& sync} denotes simulator-side execution validation and progression of the simulator state before real execution. \emph{Score-guided sampling} forms a directional prior from trajectory scores; \emph{Execution logs} support optional offline prompt refinement. A single GPT-6 Astra evaluator assesses the rendered candidates. Translation, stage-specific orientation, and endpoint gripper commands form separate control channels (Section~\ref{sec:Architecture}).}
\label{fig:antigrounding_framework}
\end{figure*}

\subsection{Simulation-Assisted Planning}
VLMPC evaluates predicted images, and Traj-VLMPC performs trajectory-based planning and evaluation~\cite{zhao2024vlmpc,trajvlmpc}. Prompting with the Future (PWTF) assesses simulated outcomes in a digital twin; SIMPACT refines actions through physics rollouts~\cite{pwtf,simpact}. AntiGrounding presents proposed trajectories rendered in the current simulator state to the VLM, rather than predicted future outcomes. Simulation also filters proposals and validates the selected segment before real execution.

\section{Method}
\label{sec:Method}
AntiGrounding takes an instruction $D$ and plans over a receding horizon in an initialized digital twin (Fig.~\ref{fig:antigrounding_framework}). Real2Sim establishes scene geometry and initial object poses. Subsequent VLM observations are multi-view renders of the evolving CoppeliaSim state, and the real robot executes simulator-validated segments. After geometric filtering, each retained candidate maintains its identity through visual assessment, simulator validation, and physical execution.

\subsection{Visual Action Representation}
\label{sec:real2sim2real}
At planning step $t$, the sampler proposes short translational end-effector motions, forming $\mathcal{T}_t=\{T_t^1,\ldots,T_t^{N_t}\}$. Linear interpolation connects each sampled target position to the current end-effector position. Ordinary motion retains the current orientation. Separate control channels supply task-specific orientation targets and gripper commands (Section~\ref{sec:Architecture}).

Proposal-level filtering checks the full interpolated segment in CoppeliaSim for non-target collisions, invalid geometric or physical states, and target-interaction or penetration constraints. Intended target contact is distinguished from collisions with surrounding objects. Fully valid segments remain intact. For a segment that violates a constraint, the ordered prefix before the first violation is retained if usable; otherwise, the proposal is discarded. The retained candidates form the feasible set $\mathcal{F}_t$.

Each candidate carries a consistent identifier and a distinct color across rendered views. Optional Set-of-Mark (SoM) annotations identify task-relevant objects~\cite{yang2023set}. Rendering exposes the retained trajectory to multimodal assessment while preserving its explicit waypoint geometry for validation and execution. Multiple projections reveal motion--scene relationships that a single view can obscure.

\subsection{Structured Multi-View Assessment}
\label{sec:VLM}
The per-view VQA prompt~\cite{sermanet2024robovqa} scores each candidate on safety, task alignment, efficiency, and physical plausibility. Safety covers collision risk and motion stability. Task alignment measures progress toward the instruction's semantic goal at the current task stage. Efficiency reflects path length and smoothness, while physical plausibility assesses whether the motion is physically reasonable in the depicted scene. A short pouring approach, for example, can score well on efficiency while leaving the cup misaligned with the bowl.

\begin{figure}[t]
\centering
\includegraphics[width=\columnwidth]{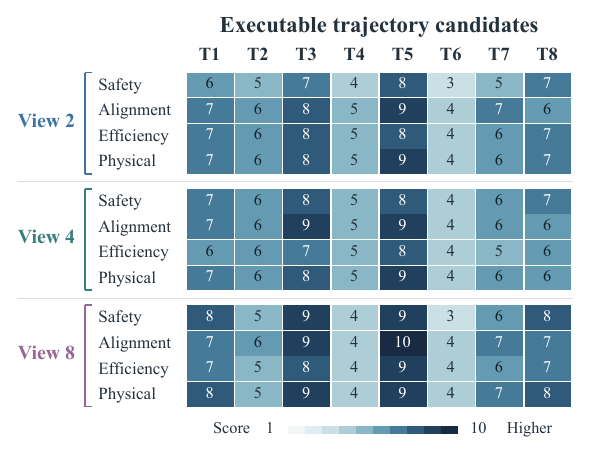}
\caption{\textbf{Multi-view criterion assessment.} Columns identify eight trajectory candidates; rows group safety, task alignment, efficiency, and physical plausibility within View 2, View 4, and View 8. Darker cells indicate higher criterion scores on a common 1--10 scale. Candidate IDs are labeled T1--T8.}
\label{fig:score_heatmap}
\end{figure}

Figure~\ref{fig:score_heatmap} shows the criterion scores for eight candidates across three views. The initial criterion weights are $(0.25,0.35,0.20,0.20)$, in the order listed above. A separate question, $q_{\mathrm{view}}$, assesses view clarity and usefulness.

For candidate $j$, criterion $k$, and view $v$, let $s_{m,v,j,k,t}$ denote the score from evaluator $m$, with $M\geq1$. The criterion weights satisfy $w_k\geq0$ and $\sum_k w_k=1$. View ranking averages the criterion-weighted scores over evaluators and the nonempty feasible candidate set:
\begin{equation}
s_{v,t}=\frac{1}{|\mathcal{F}_t|M}
\sum_{T_t^j\in\mathcal{F}_t}\sum_{m=1}^{M}\sum_{k=1}^{4}
w_k s_{m,v,j,k,t}.
\label{eq:view_ranking}
\end{equation}
Eight fixed virtual viewpoints are available, and each planning step maintains three active views $V_t$. The highest-scoring unused view replaces the lowest-scoring active view whenever its $s_{v,t}$ is higher. Replacement continues until no further improvement remains. Candidate viewpoints are scored as required for this search, but only the three retained active views enter the fusion in Eq.~\eqref{eq:aggregated_score}. Evaluating unused viewpoints also incurs query cost.

After view selection, the clarity/usefulness response $q_{v,t}$ determines the fusion weights:
\begin{equation}
\alpha_{v,t}=
\frac{q_{v,t}/(1+\lambda_C\sigma_{v,t})}
{\sum_{u\in V_t}q_{u,t}/(1+\lambda_C\sigma_{u,t})},
\label{eq:cvt}
\end{equation}
with a positive denominator. The single GPT-6 Astra configuration sets $M=1$ and $\sigma_{v,t}=0$, so weighting depends on view clarity and usefulness. An optional multi-evaluator extension penalizes cross-model disagreement through $\sigma_{v,t}$ (Appendix~\ref{app:evaluation}).

The fused trajectory score is
\begin{equation}
S_{j,t}=\sum_{v\in V_t}\alpha_{v,t}
\sum_{k=1}^{4}w_k\left(\frac{1}{M}\sum_{m=1}^{M}s_{m,v,j,k,t}\right).
\label{eq:aggregated_score}
\end{equation}
\begin{figure*}[t]
\centering
\includegraphics[width=\textwidth]{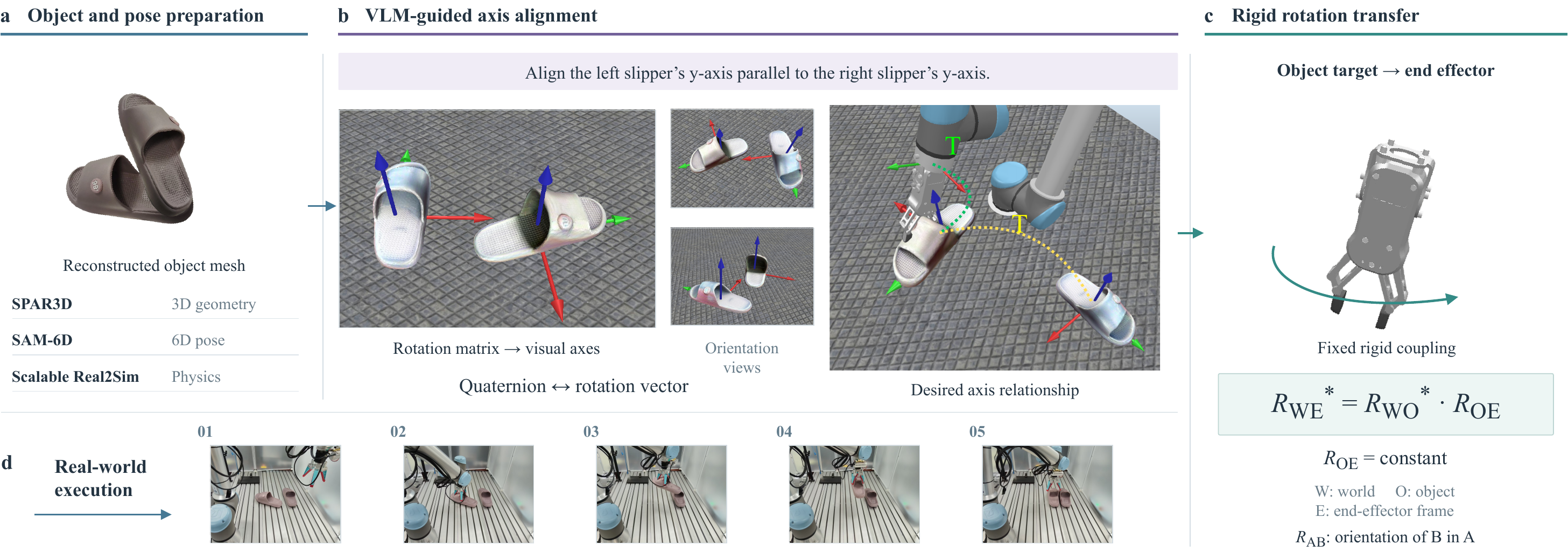}
\caption{\textbf{Object-to-end-effector orientation transfer.} The VLM infers the desired relationship between rendered object-local axes from the task instruction. Geometric conversion and the fixed object--end-effector transform map the desired object orientation to frame $E$ under a non-slipping grasp. Spherical linear interpolation (SLERP) connects the current and target orientations along the selected segment when alignment is required. Slipper Arrangement illustrates orientation-sensitive execution.}
\label{fig:pose}
\end{figure*}

The score $s_{v,t}$ ranks viewpoints, $q_{v,t}$ determines retained-view weights $\alpha_{v,t}$, and $S_{j,t}$ ranks trajectories. The highest-scoring feasible candidate proceeds to execution validation with its retained waypoint sequence.

\subsection{Evaluation-Guided Motion and Execution}
\label{sec:Sampling}
\label{sec:Architecture}
Let $p_t$ be the current end-effector position and $p^j_{\mathrm{end},t}$ the endpoint of retained candidate $T_t^j$. For nonzero translational displacement, the candidate's unit direction and the score-weighted directional prior are
\begin{equation}
\begin{aligned}
d_{j,t}&=\frac{p^j_{\mathrm{end},t}-p_t}{\|p^j_{\mathrm{end},t}-p_t\|_2},\\
g_t&=\sum_{T_t^j\in\mathcal{F}_t}S_{j,t}d_{j,t},\qquad
\hat g_t=\frac{g_t}{\|g_t\|_2}.
\end{aligned}
\label{eq:directional_prior}
\end{equation}
For $g_t\ne0$, subsequent proposals are sampled around $\hat g_t$ within the active angular dispersion and sampling radius. All feasible candidates contribute through their aggregate scores $S_{j,t}$.

Within each semantic stage, the candidate count contracts from eight to three, the sampling radius from 0.25 to 0.10\,m, and angular dispersion from $90^\circ$ to $30^\circ$. This coarse-to-fine schedule shifts proposal search from broad exploration to local refinement. Appendix~\ref{app:implementation} specifies the $R$ and $\theta$ schedules.

For orientation-sensitive interaction, the VLM infers the required axis relationship from the instruction and rendered object-local $X/Y/Z$ axes (Fig.~\ref{fig:pose}). Deterministic geometric conversion produces the desired object orientation, which rigid coupling transfers to the end-effector:
\begin{equation}
{}^{W}R_E^*={}^{W}R_O^*\,{}^{O}R_E.
\label{eq:rigid}
\end{equation}
Here $W$, $O$, and $E$ denote the world, object, and end-effector frames, and $ {}^{A}R_B$ maps coordinates from $B$ to $A$. The transform $ {}^{O}R_E$ remains fixed under a non-slipping grasp. Unconstrained rotation follows the existing grasp/orientation convention. When a stage-specific target is supplied, SLERP interpolates from the current orientation to $ {}^{W}R_E^*$ along the selected segment. Ordinary motion retains the current orientation.

The VLM infers semantic stages from the instruction, rendered scene, and execution progress. Its structured response supplies global gripper open/close, stage-transition, and task-completion signals.

The selected segment undergoes a second simulator check through execution validation, including any required orientation interpolation. A failed rollout is rejected without committing it to the planning state or actuating the real robot; planning resumes from the pre-validation state. Successful validation retains the simulated post-action state and permits the UR5e to execute the same interpolated waypoint sequence. In both simulation and real execution, the gripper opens or closes according to its signal at the interaction endpoint, after motion and required alignment are complete.

A stage-transition signal takes effect in the next planning round, after segment execution and the required gripper action. It resets the directional prior, candidate count, sampling radius, and angular dispersion. A completion signal terminates the episode after the final segment and required gripper action. The real robot's low-level waypoint tracking operates separately from this high-level planning loop.

Optional offline prompt refinement analyzes execution logs to revise criterion weights, question wording, or task-specific evaluation criteria while keeping VLM weights fixed (Appendix~\ref{sec:Optimization}).

\section{Experiments}
\label{Experiments}
The eight tasks span placement, obstacle-aware motion, multi-stage manipulation, and semantic constraints (Fig.~\ref{fig:head}). Experiments use a UR5e with a three-finger flexible gripper. Appendix~\ref{app:protocol} and Fig.~\ref{fig:equipment} describe the platform and task protocol.

\subsection{Real-World Manipulation Performance}
\label{sec:Performance}
\begin{table*}[t]
\centering
\caption{Eight-task real-world manipulation comparison. Entries are successes out of ten per task; overall success is computed over 80 episodes.}
\label{tab:table1}
\small
\begin{tabular}{lccccccccc}
\toprule
Method & Tape & Sponge & Maze & Drawer & Water & Slipper & Waste & Shape & \shortstack{Overall\\(\%)}\\
\midrule
\multicolumn{10}{l}{\textbf{Robot-policy / VLA deployment references}}\\
OpenVLA & 0/10 & 0/10 & 2/10 & 1/10 & 0/10 & 0/10 & 1/10 & 1/10 & 6.25\\
OpenVLA-OFT & 2/10 & 2/10 & 4/10 & 3/10 & 1/10 & 1/10 & 2/10 & 1/10 & 20.00\\
$\pi_0$ & 1/10 & 1/10 & 3/10 & 2/10 & 1/10 & 1/10 & 1/10 & 1/10 & 13.75\\
$\pi_{0.5}$ & \PiTape/10 & \PiSponge/10 & \PiMaze/10 & \PiDrawer/10 & \PiWater/10 & \PiSlipper/10 & \PiWaste/10 & \PiShape/10 & \PiOverall\\
\midrule
\multicolumn{10}{l}{\textbf{Reasoning / planning methods}}\\
Code as Policies & 1/10 & 2/10 & 1/10 & 1/10 & 0/10 & 0/10 & 0/10 & 0/10 & 6.25\\
VoxPoser & 3/10 & 2/10 & 3/10 & 4/10 & 0/10 & 0/10 & 2/10 & 1/10 & 18.75\\
ReKep & 5/10 & 6/10 & 3/10 & 4/10 & 6/10 & 5/10 & 2/10 & 2/10 & 41.25\\
PIVOT-style$^{\dagger}$ & \PivotTape/10 & \PivotSponge/10 & \PivotMaze/10 & \PivotDrawer/10 & \PivotWater/10 & \PivotSlipper/10 & \PivotWaste/10 & \PivotShape/10 & \PivotOverall\\
\textbf{AntiGrounding}$^{\dagger}$ & \textbf{\AstraTape/10} & \textbf{\AstraSponge/10} & \textbf{\AstraMaze/10} & \textbf{\AstraDrawer/10} & \textbf{\AstraWater/10} & \textbf{\AstraSlipper/10} & \textbf{\AstraWaste/10} & \textbf{\AstraShape/10} & \textbf{\AstraOverall}\\
\bottomrule
\end{tabular}
\par\smallskip
\begin{minipage}{\textwidth}
\footnotesize
The four VLA baselines use pretrained deployment policies with the robot-side deployment and action-interface adaptation required for the UR5e. They are evaluated zero-shot on the eight tasks, without task-specific fine-tuning using data from these tasks. OpenVLA-OFT denotes Optimized Fine-Tuning and uses the released OFT checkpoint. $^{\dagger}$ Both methods use the same single GPT-6 Astra evaluator; PIVOT-style denotes the adapted visual proposal-selection baseline. Bold denotes the best per-task and overall results.
\end{minipage}
\end{table*}

\begin{figure*}[t]
\centering
\includegraphics[width=\textwidth]{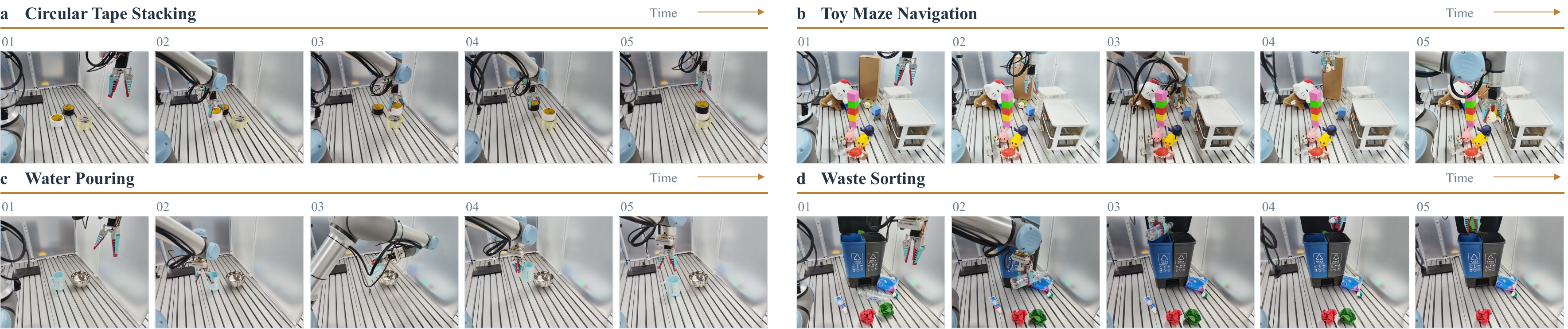}
\caption{\textbf{Representative real-world execution sequences.} (a) Circular Tape Stacking: ring placement and alignment. (b) Toy Maze Navigation: obstacle-aware navigation. (c) Water Pouring: orientation-sensitive multi-stage manipulation. (d) Waste Sorting: semantic sorting. Labels 01--05 mark representative snapshots in execution order.}
\label{fig:real_task}
\end{figure*}

We evaluated nine methods over ten real-world episodes per task (Table~\ref{tab:table1}). All methods shared the task protocol and reset ranges, with initial conditions sampled independently across methods. Resets varied initial object and robot poses, viewpoint, and scene layout where applicable. Success was binary completion of the task-specific objective.

Robot-policy deployment references comprise OpenVLA~\cite{kim2024openvla}, OpenVLA-OFT~\cite{oft}, $\pi_0$~\cite{black2024pi_0}, and $\pi_{0.5}$~\cite{intelligence2025pi_0.5}. The reasoning and planning baselines are Code as Policies, VoxPoser, ReKep~\cite{codeaspolicy,voxposer,huang2024rekep}, and PIVOT-style~\cite{pivot}.

The PIVOT-style adaptation shares AntiGrounding's candidate generation, rendering, feasibility filtering, simulator validation, and execution infrastructure, including the $N/R/\theta$ search controls. At each planning round, GPT-6 Astra receives the instruction and ID/color-annotated proposals and directly returns a candidate ID, replacing structured multi-view scoring. The selected segment undergoes simulator validation and execution; its direction guides the next proposal set instead of the score-weighted prior in Eq.~\eqref{eq:directional_prior}. Replanning follows the VLM stage-transition and completion signals.

AntiGrounding achieves \AstraTotal/80 successes (\AstraOverall\%), compared with \PiTotal/80 (\PiOverall\%) for $\pi_{0.5}$ and \PivotTotal/80 (\PivotOverall\%) for PIVOT-style. These correspond to differences of 21.25 and 23.75 percentage points under the reported deployment protocol. AntiGrounding completes six to eight episodes per task.

Figure~\ref{fig:real_task} illustrates representative executions. In Water Pouring (c), the robot brings the green cup above the bowl and tilts it to transfer water. Translational candidates determine the approach while the end-effector retains its orientation. During pouring, the VLM assesses the desired cup-axis relationship. Rigid object--end-effector coupling maps the resulting object-orientation target to the end-effector, and SLERP interpolates the rotation along the selected segment. The gripper holds the cup during transport and pouring; grasp and release commands take effect at interaction endpoints.

\subsection{Component Ablations}
\label{sec:Ablation}
Table~\ref{tab:ablation_slim} reports four component ablations with the same single GPT-6 Astra evaluator, using 30 real-world trials per condition and task. All conditions shared the task protocols and reset ranges, with trials sampled independently across conditions. These trials were independent of the ten-episode evaluations in Table~\ref{tab:table1}. Full AntiGrounding succeeds in 63.33--83.33\% of trials across the four tasks.
\begin{table*}[t]
\centering
\caption{Component ablations with a single GPT-6 Astra evaluator. Each entry reports successes out of 30 real-world trials and the corresponding percentage; higher is better.}
\label{tab:ablation_slim}
\small
\begin{tabular}{lcccc}
\toprule
Variant & Circular Tape Stacking & Toy Maze Navigation & Water Pouring & Waste Sorting\\
\midrule
\textbf{AntiGrounding} & \textbf{\AblFullTape} & \textbf{\AblFullMaze} & \textbf{\AblFullWater} & \textbf{\AblFullWaste}\\
AG-SVT & \AblSvtTape & \AblSvtMaze & \AblSvtWater & \AblSvtWaste\\
AG-MVECW & \AblMvecwTape & \AblMvecwMaze & \AblMvecwWater & \AblMvecwWaste\\
AG-VGTG & \AblVgtgTape & \AblVgtgMaze & \AblVgtgWater & \AblVgtgWaste\\
AG-RCRM & \AblRcrmTape & \AblRcrmMaze & \AblRcrmWater & \AblRcrmWaste\\
\bottomrule
\end{tabular}
\end{table*}

\textbf{Assessment organization.}
AG-SVT replaces the four-criterion decomposition with a single holistic trajectory score from the same evaluator, leaving the remaining assessment and execution pipeline unchanged. Success falls to 30.00--53.33\%. AG-MVECW retains one predefined canonical viewpoint throughout each episode, jointly removing view diversity, adaptive replacement, and cross-view fusion. Success falls to 30.00--43.33\%. Both ablations reduce success across all four tasks.

\textbf{Proposal search and orientation control.}
AG-VGTG removes score-weighted directional guidance and the coarse-to-fine $N/R/\theta$ schedule, sampling candidate motions uniformly within the original search space. Success falls to 3.33--10.00\%. AG-RCRM disables orientation-sensitive rigid transfer, so the end-effector retains its current orientation during translation. Success reaches 3.33--20.00\%. Removing proposal refinement or rigid orientation transfer produces larger degradations than either assessment ablation.

\subsection{Evaluator Sensitivity}
\label{sec:Evaluator}
\begin{table}[t]
\centering
\caption{Evaluator sensitivity across eight tasks using the same robot-side AntiGrounding pipeline.}
\label{tab:evaluator}
\small
\begin{tabular}{lcc}
\toprule
Evaluator & Successes & Overall (\%)\\
\midrule
Gemini 2.5 Pro & 43/80 & 53.75\\
HunYuan Vision & 40/80 & 50.00\\
GLM-4v-Plus & 41/80 & 51.25\\
Gemini + GLM & 48/80 & 60.00\\
Gemini + GLM + HunYuan & 52/80 & 65.00\\
\midrule
\textbf{GPT-6 Astra (single)} & \textbf{\AstraEvalCount/80} & \textbf{\AstraEvalOverall}\\
\bottomrule
\end{tabular}
\par\smallskip
\begin{minipage}{\columnwidth}
\footnotesize
The first five configurations aggregate responses from all configured evaluators under the complete-response protocol; the GPT-6 Astra row uses the Table~\ref{tab:table1} condition.
\end{minipage}
\end{table}

Table~\ref{tab:evaluator} compares evaluator configurations with the robot-side AntiGrounding pipeline held fixed. Single GPT-6 Astra records \AstraEvalCount/80 successes (\AstraEvalOverall\%), compared with 52/80 (65.00\%) for the three-model ensemble. These configurations require no retraining of the robot-side method.

\subsection{Failure Modes and Operating Limits}
\label{sec:Failures}
VLM assessment errors include incorrect trajectory rankings, cross-view judgments, and axis-alignment judgments. Appendix~\ref{app:failures} summarizes the categorized failures.

Performance depends on candidate coverage and the fidelity of the initialized digital twin. Slip, friction mismatch, deformation, contact-model error, inaccurate physical parameters, and calibration error can cause real execution to diverge from simulation. High-level planning continues from the simulator state without fresh real-image observations or per-segment scene re-registration. Virtual-view changes alter the projection of that state, while physical-robot feedback supports low-level waypoint tracking. Scene initialization and online VLM decision latency are separate costs; assessments of unused viewpoints contribute to online latency.

\section{Conclusion}
\label{Conclusion and Limitations}
We presented AntiGrounding, a visual action-selection framework built around a dual geometric--visual trajectory interface. Each retained candidate serves as both a geometric control object and a visual action representation through VLM assessment, simulator validation, and real execution. The eight-task study demonstrates manipulation with a single GPT-6 Astra evaluator. The evaluator study compares multimodal backends with the robot-side method held fixed. CoppeliaSim maintains the primary planning state, and the real robot follows validated trajectory segments. Performance remains constrained by digital-twin fidelity, candidate quality, and physical interaction.

\appendix
\section{Implementation and Evaluation Details}
\label{app:details}

\begin{figure*}[t]
\centering
\includegraphics[width=\textwidth]{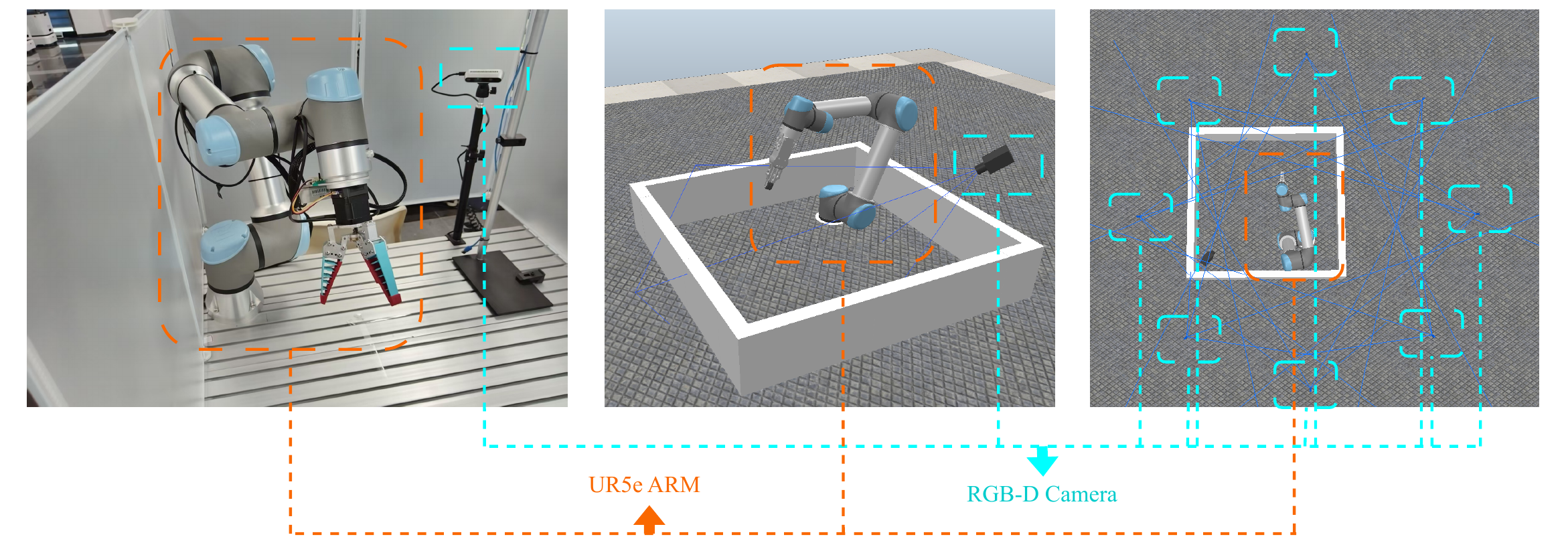}
\caption{\textbf{Experimental platform and virtual viewpoints.} Left: the real UR5e and RGB-D camera for initial scene construction and pose alignment. Center: the corresponding simulator setup. Right: the virtual-viewpoint pool for rendering the primary planning state. The real robot executes simulator-validated waypoint sequences.}
\label{fig:equipment}
\end{figure*}

\subsection{Scene, Sampling, and Assessment}
\label{app:implementation}
\label{app:evaluation}
SPAR3D reconstructs object geometry from single RGB images~\cite{huang2025spar3d}. Articulated structures and regular objects, such as drawers and blocks, use specified models. SAM-6D estimates the initial object poses~\cite{lin2024sam}. Interaction-sensitive tasks can additionally incorporate mass, center-of-mass, and inertia estimates from Scalable Real2Sim~\cite{pfaff2025scalable} or physical-parameter identification. Tasks dominated by geometry and collision constraints primarily rely on geometry and pose. Meshes are reused across episodes with the same objects, so resets mainly update the initial poses.

Linear interpolation generates translational waypoints. The end-effector retains its orientation unless a stage-specific target activates SLERP, and the independent gripper command takes effect after endpoint arrival. Whole-segment proposal filtering and selected-action execution validation precede real actuation.

The annealing schedules use semantic-stage step $t'$:
\begin{equation}
R_{t'}=R_{\min}+(R_0-R_{\min})e^{-\lambda_R t'},
\label{eq:radius_decay}
\end{equation}
\begin{equation}
\theta_{t'}=\theta_{\min}+(\theta_0-\theta_{\min})e^{-\lambda_\theta t'}.
\label{eq:theta_decay}
\end{equation}
The settings are $R_0=0.25$\,m, $R_{\min}=0.10$\,m, $\lambda_R=0.693$, $\theta_0=90^\circ$, $\theta_{\min}=30^\circ$, and $\lambda_\theta=0.712$. Candidate count $N$ also contracts from eight to three. Stage transitions reset $N$, $R$, $\theta$, and directional guidance after the current segment and gripper action.

Criterion scores range from 1 to 10, with higher scores preferred. A separate question rates view clarity and usefulness from 0 to 10. Candidate viewpoints are evaluated as required for ranking and replacement. The assessment prompt presents the three retained active views in a horizontally concatenated layout and supplies candidate ID--color mappings. It maps the left, middle, and right images to their view IDs. Responses group the four criterion scores by view ID and trajectory ID, with a separate clarity/usefulness score for each view. Gripper, stage-transition, and completion signals are global response fields.

For the optional multi-evaluator extension, disagreement is
\begin{equation}
\sigma_{v,t}=
\sqrt{\frac{1}{|\mathcal{F}_t|}
\sum_{T_t^j\in\mathcal{F}_t}
\mathrm{Var}_{m}\!\left(\sum_{k=1}^{4}w_k s_{m,v,j,k,t}\right)}.
\label{eq:sigma}
\end{equation}
Variance is computed across evaluators in a common response set, and the disagreement penalty uses $\lambda_C=0.5$. The single-evaluator configuration sets $\sigma_{v,t}=0$.

\begin{figure}[b]
\centering
\includegraphics[width=\columnwidth]{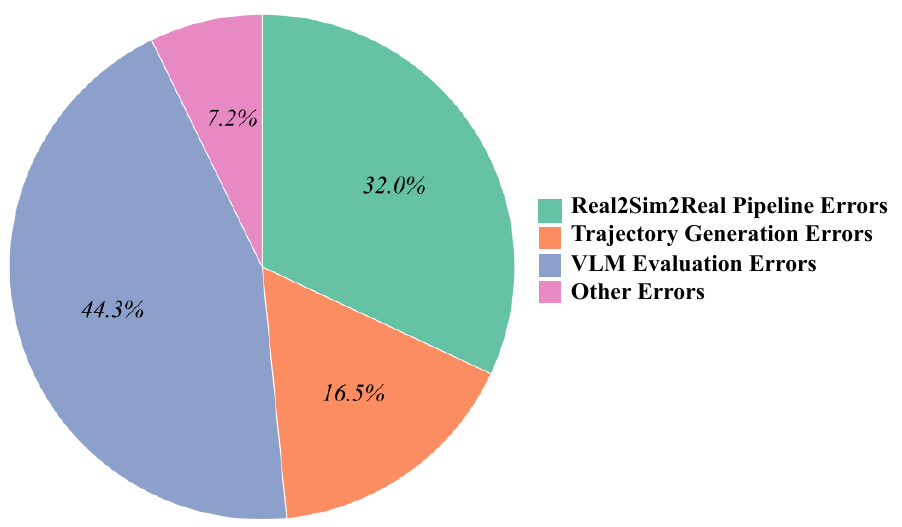}
\caption{Categorized failures in a separate failure-analysis cohort. Percentages denote shares of categorized failures.}
\label{fig:pie}
\end{figure}

\subsection{Real-World Protocol}
\label{app:protocol}
The platform comprises a UR5e with a three-finger flexible gripper on a $1.0\times0.5\times0.70$\,m table (Fig.~\ref{fig:equipment}). A fixed side-view Intel RealSense D435i, configured at 30\,Hz, supports initial Real2Sim construction and pose alignment. Subsequent VLM assessments operate on simulator renders. CoppeliaSim 4.9.0 with Bullet 2.7.8 provides the primary scene state and motion checks. The simIK plugin computes inverse-kinematics solutions for end-effector waypoint tracking. The real robot follows validated waypoint sequences through Real-Time Data Exchange (RTDE) position control at 125\,Hz, independently of the VLM decision frequency.

The VLM task-completion signal terminates control after the final segment and gripper action. Experimental success is scored from the resulting real-world outcome under the task-specific objective.

Circular Tape Stacking requires three rings to be stacked vertically with concentric alignment. Sponge Stacking places one sponge on another with aligned edges. Toy Maze Navigation retrieves a toy turtle through obstacles; Drawer Retrieval extracts a screwdriver from a partially opened drawer while avoiding its edges. Water Pouring transfers water from a green cup to a steel bowl. Slipper Arrangement places slippers side by side in left--right order with aligned toes. Waste Sorting places a tissue pack in the task-designated recyclable bin and a toothpaste tube in the non-recyclable bin. Shape Matching places pieces into matching slots with the required pose.

\subsection{Optional Offline Prompt Refinement}
\label{sec:Optimization}
A meta-VLM examines execution logs containing candidates, rendered images, raw responses, parsed scores, control signals, and observed outcomes. Prediction--outcome mismatches guide revisions to criterion weights, question wording, or task-specific evaluation criteria. The revised prompt replaces the active evaluator prompt while VLM weights remain fixed. Observed outcomes are available only for executed trajectories.

The evaluations in Tables~\ref{tab:table1}--\ref{tab:evaluator} use a fixed evaluator prompt. Optional offline prompt refinement is not applied during the reported episodes.

\subsection{Categorized Failures}
\label{app:failures}
Figure~\ref{fig:pie} summarizes a separate failure-analysis cohort. VLM assessment accounts for the largest share of categorized failures, followed by scene-building errors. The remaining categories concern trajectory generation and execution.

\bibliographystyle{IEEEtran}
\bibliography{main}
\end{document}